\def\year{2021}\relax
\documentclass[letterpaper]{article} 
\usepackage{aaai21}  
\usepackage{times}  
\usepackage{helvet} 
\usepackage{courier}  
\usepackage[hyphens]{url}  
\usepackage{graphicx} 
\usepackage{natbib}  
\usepackage{caption} 
\usepackage[utf8]{inputenc} 
\usepackage{booktabs}       
\usepackage{amsfonts}       
\usepackage{nicefrac}       
\usepackage{microtype}      

\usepackage{subfig}
\usepackage{algorithmicx}
\usepackage{algpseudocode}
\usepackage{amsmath}
\usepackage{textcomp}
\usepackage{xcolor}
\usepackage[ruled,vlined]{algorithm2e}
\usepackage{amssymb}
\usepackage{makecell}
\usepackage{multirow}

\SetCommentSty{mycommfont}

\usepackage{xspace}

\usepackage{subfig}
\usepackage{enumitem}

\newcommand{\projectname}{RT3D\xspace}

\definecolor{Xue_color}{rgb}{0.858, 0.188, 0.478}
\usepackage[switch]{lineno}

\title{\projectname: Achieving Real-Time Execution of 3D Convolutional Neural Networks on Mobile Devices}
\author{
    Wei Niu\textsuperscript{\rm 1}\thanks{These authors contributed equally.},
    Mengshu Sun\textsuperscript{\rm 2}\text{$^*$}, Zhengang Li\textsuperscript{\rm 2}\text{$^*$}, Jou-An Chen\textsuperscript{\rm 3}, Jiexiong Guan\textsuperscript{\rm 1},\\
    Xipeng Shen\textsuperscript{\rm 3}, Yanzhi Wang\textsuperscript{\rm 2}, Sijia Liu\textsuperscript{\rm 4}, Xue Lin\textsuperscript{\rm 2}, Bin Ren\textsuperscript{\rm 1} \\
}
\affiliations{

    \textsuperscript{\rm 1} William \& Mary, 
    \textsuperscript{\rm 2} Northeastern University, 
    \textsuperscript{\rm 3} North Carolina State University,
    \textsuperscript{\rm 4} MIT-IBM Watson AI Lab\\

    \textsuperscript{\rm 1}\{wniu, jguan\}@email.wm.edu,
    \textsuperscript{\rm 2}\{sun.meng, li.zhen, yanz.wang, xue.lin\}@northeastern.edu,\\
    \textsuperscript{\rm 3}\{jchen73, xshen5\}@ncsu.edu,
    \textsuperscript{\rm 4}sijia.liu@ibm.com,
    \textsuperscript{\rm 1}bren@cs.wm.edu\\

}

\begin{document}

\maketitle

\begin{abstract}
Mobile devices are becoming an important carrier for deep learning tasks, as they are being equipped with powerful, high-end mobile CPUs and GPUs. However, it is still a challenging task to execute 3D Convolutional Neural Networks (CNNs) targeting for real-time performance, besides high inference accuracy. The reason is more complex model structure and higher model dimensionality overwhelm the available computation/storage resources on mobile devices. A natural way may be turning to deep learning weight pruning techniques. However, the direct generalization of existing 2D CNN weight pruning methods to 3D CNNs is not ideal for fully exploiting mobile parallelism while achieving high inference accuracy.

This paper proposes RT3D, a model compression and mobile acceleration framework for 3D CNNs, seamlessly integrating neural network weight pruning and compiler code generation techniques. We propose and investigate two structured sparsity schemes i.e., the vanilla structured sparsity and kernel group structured (KGS) sparsity that are mobile acceleration friendly. The vanilla sparsity removes whole kernel groups, while KGS sparsity is a more fine-grained structured sparsity that enjoys higher flexibility while exploiting full on-device parallelism. We propose a reweighted regularization pruning algorithm to achieve the proposed sparsity schemes. The inference time speedup due to sparsity is approaching the pruning rate of the whole model FLOPs (floating point operations). RT3D  demonstrates up to $29.1\times$ speedup in end-to-end inference time comparing with current mobile frameworks supporting 3D CNNs, with moderate $1\%\sim1.5\%$ accuracy loss. The end-to-end inference time for 16 video frames could be within 150 ms, when executing representative C3D and R(2+1)D models on a cellphone. For the first time, real-time execution of 3D CNNs is achieved on off-the-shelf mobiles.

\end{abstract}  

\section{Introduction}

Since the Convolutional Neural Networks (CNNs) were exemplified by the performance improvements obtained by AlexNet~\cite{krizhevsky2012imagenet} in 2012, neural network based computer vision has achieved superhuman performance.
Mobile devices are becoming an important carrier for deep learning tasks.
However, real-time execution is the most critical requirement given computation/storage resource constraints on mobiles for deep learning tasks.

Recently, many efforts~\cite{han2016mcdnn,yao2017deepsense, huynh2017deepmon,chen2018tvm,Ali-MNN,TensorFlow-Lite,Pytorch-Mobile} aim to accelerate CNN execution on off-the-shelf mobile devices and some of them achieve significant advancements. 
However, most of these optimizations focus on traditional 2D CNNs in the image domain.
On the other hand, 3D CNNs have been proposed for video domain tasks such as video classification, and action recognition/detection~\cite{ji20123d,wang2017winograd,carreira2017quo,qiu2017learning, kopuklu2019resource} 
It is still an open problem to execute 3D CNNs on off-the-shelf mobile devices targeting for real-time performance.  
For example, C3D~\cite{tran2015learning}, a mainstream 3D CNN takes over 2.5 seconds 
to complete the inference (of 16 frames) on a representative mobile CPU (Kryo 585 in Qualcomm Snapdragon platform) {with Pytorch Mobile \cite{Pytorch-Mobile}}, which is clearly far from real-time execution.\footnote{Real-time performance requires to compute 30 frames/second according to state-of-the-art industry standard.} 
The extra dimension in 3D convolution (CONV) significantly increases storage size and computation workload comparing with 2D CONV.\footnote{2D CONV is a special case of 3D CONV with the temporal dimension size equal to 1.} The large memory footprint of 3D CNN models often exceeds the on-chip cache size of off-the-shelf mobile devices. As a result, 3D CNNs are currently supported only by very few mobile acceleration frameworks i.e., {PyTorch Mobile \cite{Pytorch-Mobile} and Alibaba MNN \cite{Ali-MNN}} with relatively low {computation} efficiency, let alone real-time execution performance. 

A natural way to bridge the gap is to turn to \emph{model compression} techniques, particularly \emph{weight pruning}~\cite{wen2016learning,han2015deep,guo2016dynamic,dong2019network,zhuang2018discrimination,yu2018nisp,he2019filter} which has demonstrated its efficacy on accelerating 2D CNN executions. Nevertheless, generalizing weight pruning methods from 2D to 3D CNNs is more than a straightforward task owing to {the higher dimensionality of weight tensors and thus the larger search space of weight pruning}. It is especially challenging to derive the best-suited weight pruning method in order to achieve real-time performance on off-the-shelf mobile devices. Two fundamental problems need to be solved: the \emph{sparsity scheme} and the \emph{pruning algorithm}. The former refers to the \emph{regularity} in pruning, i.e., the specific structural {characteristics} of CNNs after pruning. The two representative cases for 2D CNNs are the most flexible, irregular pruning scheme that can prune arbitrary weights \cite{han2015deep,guo2016dynamic}, and the {computing platform}-friendly 
filter/channel pruning scheme that prunes whole filters/channels \cite{wen2016learning,he2019filter,yu2018nisp}. The latter refers to the appropriate algorithm to determine the target weights to remove and train the remaining, non-zero weights. For 2D CNNs, there is also rich literature in heuristic pruning \cite{han2015deep,guo2016dynamic,dong2019network} or regularization-based pruning algorithms \cite{wen2016learning,yu2018nisp,he2019filter}.

{This work develops RT3D framework, including the derivation of best-suited (structured) sparsity scheme and pruning algorithm of 3D CNNs, and the design of the associated compiler-aided acceleration, for off-the-shelf mobile devices}.
We propose and investigate \emph{two structured sparsity schemes} that are highly mobile acceleration friendly. The first vanilla sparsity scheme achieves sparsity by removing kernel groups in 3D CONV layers. It can achieve straightforward acceleration for on-device inference with the aid of compiler code generation, but it suffers from relatively high accuracy loss as whole kernel groups are pruned. The second, more optimized one is the \emph{kernel group structured} (KGS) sparsity scheme. It is a more fine-grained structured sparsity that enjoys higher flexibility, and will result in a higher accuracy under the same pruning rate. Moreover, it is important to note that the KGS sparsity scheme is beyond a mere tradeoff of accuracy and mobile performance. In fact, with proper support of compiler code generation, the KGS sparsity can \emph{achieve almost the same mobile acceleration (e.g., in frames/second) as the vanilla sparsity}, under the same pruning rate. This is owing to the delicate design of KGS sparsity to match the parallelization mechanism in compiler-assisted mobile acceleration, such that the full on-device parallelism can be exploited. 

We further present \emph{three pruning algorithms} to achieve the proposed structured sparsity schemes for 3D CNNs. The first two, i.e., the heuristic algorithm and regularization-based algorithm, are natural generalization from state-of-the-art algorithms on 2D CNN weight pruning. However, they are either greedy algorithm, or suffer from the limitation that all weights will be equally penalized even after convergence of the pruning process. Both result in potential accuracy loss. To overcome these shortcomings, we propose a novel \emph{reweighted regularization pruning} algorithm. The basic idea is to systematically and dynamically reweight the penalties, reducing the penalties on weights with large magnitudes (which are likely to be more critical), and increasing the penalties on weights with smaller magnitudes. It possesses other advantages, such as not introducing additional hyperparameters, and being flexible for either parameter reduction or FLOPs (floating-point operations) reduction, etc.

Seamlessly integrated with above innovations, \projectname also develops the \emph{first} end-to-end, compiler-assisted acceleration framework of 3D CNNs on both mobile CPUs and GPUs (the few prior work are limited to mobile CPUs), and also the \emph{first} to support different structured sparsity schemes. RT3D achieves up to $29.1\times$ speedup in end-to-end inference time comparing with current mobile frameworks supporting 3D CNNs, with moderate 1\%-1.5\% accuracy loss, on representative CNNs (C3D, R(2+1)D, S3D). The end-to-end inference time for 16 video frames could be within 150 ms. 

A brief contribution summary is: (a) sparsity schemes for 3D CNNs which are both flexible and mobile acceleration friendly, (b) highly effective pruning algorithm to achieve such sparsity schemes, (c) compiler-assisted mobile acceleration framework, and (d) for the first time, real-time performance of 3D CNNs can be achieved on off-the-shelf mobile devices using a pure software solution.

\section{Related Work}

\subsection{Weight Pruning for 2D CNNs}\label{sec:relatedwork}

The rich literature in weight pruning for 2D CNNs can be categorized into \emph{heuristic pruning algorithms} and \emph{regularization-based pruning algorithms}. The former starts from the early work on irregular, unstructured weight pruning where arbitrary weights can be pruned. \cite{han2015deep} adopts an iterative algorithm to eliminate weights with small magnitude and perform retraining to regain accuracy. \cite{guo2016dynamic} incorporates connection splicing into the pruning process to dynamically recover the pruned connections that are found to be important. Later, heuristic pruning algorithms have been generalized to the more hardware-friendly structured sparsity schemes. In~\cite{dong2019network}, Transformable Architecture Search (TAS) is adopted to realize the pruned network and knowledge is transferred from the unpruned network to the pruned version. The work~\cite{luo2017thinet} leverages a greedy algorithm to guide the pruning of the current layer with input information of the next layer. The work \cite{yu2018nisp} defines a ``neuron importance score" and propagates this score to conduct the weight pruning process.

Regularization-based pruning algorithms, on the other hand, are more mathematics-oriented and have the unique advantage for dealing with structured pruning problems through group Lasso regularization \cite{yuan2006model,liu2018rethinking}. Early work \cite{wen2016learning,he2017channel} incorporate $\ell_1$ or $\ell_2$ regularization in loss function to solve filter/channel pruning problems.  However, there is also one limitation of the direct application of regularization terms -- all weights will be penalized equally even after pruning convergence, resulting in potential accuracy loss. A number of subsequent work are dedicated to making the regularization penalty a dynamic and "soft" term. The method in~\cite{he2018soft} selects filters based on $\ell_2$ norm and updates the filters that have been previously pruned. \cite{zhang2018systematic,li2019compressing} incorporate the advanced optimization solution framework ADMM (Alternating Direction Methods of Multipliers) to achieve dynamic regularization penalty, thereby improving accuracy. \cite{he2019filter} proposes to adopt Geometric Median, a classic robust estimator of centrality for data in Euclidean spaces. A common limitation of these improved versions is that the pruning rate for each layer needs to be manually set, which is difficult to derive in prior.


\subsection{Mobile Acceleration Frameworks of CNNs}


TVM~\cite{chen2018tvm},
TFLite~\cite{TensorFlow-Lite}, Alibaba Mobile Neural Network (MNN)~\cite{Ali-MNN} and PyTorch Mobile (PyTorch)~\cite{Pytorch-Mobile} are representative compiler-assisted deep learning acceleration frameworks on mobile devices. They mainly focus on end-to-end acceleration for 2D CNNs. Only MNN and PyTorch support 3D CONV on mobile CPUs (no mobile GPU support); while other popular ones (like TVM and TFLite) do not support 3D CONV computation. To the best of our knowledge, our \projectname is the {\bf first} end-to-end deep learning acceleration framework for 3D CNNs on both mobile CPUs and GPUs. More than that, it is also the {\bf first} to support the acceleration of various sparsity schemes of 3D CNNs. 
Moreover, several hardware solutions for 3D CNN acceleration have been proposed, e.g.~\cite{hegde2018morph,shen2018towards,chen20193d}. Different from these valuable solutions that require special hardware design, \projectname employs a {\bf pure software} solution on off-the-shelf mobile devices that is more cost-effective.


\section{Structured Sparsity Schemes for 3D CNNs}

This section proposes two structured sparsity schemes of 3D CNNs. 
We focus on the most computationally intensive convolutional (CONV) layers of 3D CNNs.
Let ${\mathbf{W}}_l\in\mathbb{R}^{M\times N\times K_h\times K_w\times K_d}$ denote the 5-dimensional weight tensor of the $l$-th CONV layer of a 3D CNN, where $M$ is the number of filters; $N$ is the number of input channels; $K_w$, $K_h$, and $K_d$ are the width, height, and depth, respectively, of the 3D CONV kernels.
Different from the 2D CONV kernel, the 3D CONV kernel has an additional dimension on the kernel depth, making ${\mathbf{W}}_l$ a 5-dimensional tensor.

\begin{figure*}[hptb]
  \centering
  \includegraphics[width=1 \textwidth]{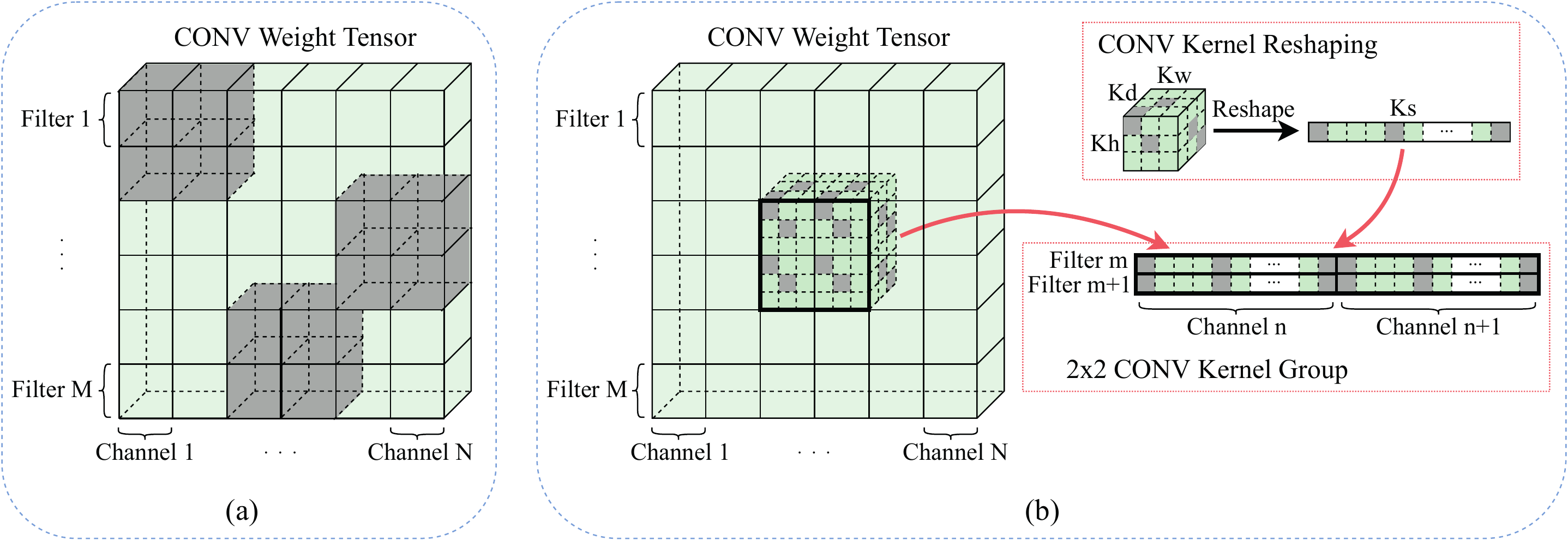}
  \caption{Proposed structured sparsity schemes: (a) The Vanilla Structured Sparsity. (b) The Kernel Group Structured (KGS) Sparsity for 3D CNNs. A CONV weight tensor is first split into multiple \emph{kernel groups}, each consisting of $g_M\times g_N$ ($2\times 2$ in the figure) 3D kernels. Within the same kernel group, kernels are pruned at the same locations (marked by the grey entries).}
  \label{fig:KGSSparsity}
\end{figure*}

Figure~\ref{fig:KGSSparsity} demonstrates the proposed two structured sparsity schemes for 3D CNNs: \emph{Vanilla Structured Sparsity Scheme} and \emph{Kernel Group Structured (KGS) Sparsity Scheme}.
The weight tensor ${\mathbf{W}}_l$ is first partitioned into \emph{groups of kernels} along the filter and input channel dimensions.
Each kernel group consists of $g_M\times g_N$ ($2\times 2$ in Figure \ref{fig:KGSSparsity}) 3D kernels.
The Vanilla sparsity scheme is shown in Figure \ref{fig:KGSSparsity} (a), where whole kernel groups are determined to be pruned or not.
On the other hand, our proposed KGS sparsity scheme as shown in Figure \ref{fig:KGSSparsity} (b) is that for the kernels in the same group, weights are pruned at the same locations. 
This is illustrated better on the right of Figure \ref{fig:KGSSparsity} (b), where 3D kernels are reshaped into vectors with $K_s = K_h \times K_w \times K_d$ weights. 
Consider the $g_M \times g_N$ kernels in a group, i.e., kernels ${\mathbf{W}}_l(m:m+g_M-1,n:n+g_N-1,:,:,:)$. Weights at the same location in these kernels i.e., ${\mathbf{W}}_l(m:m+g_M-1,n:n+g_N-1,h,w,d)$ are determined to be pruned or not together, where $(:,:,h,w,d)$ describes the same location (coordinate) in kernels.

The Vanilla sparsity scheme is a relatively straightforward generalization from structured sparsity schemes \cite{wen2016learning,liu2017learning,luo2017thinet} for 2D CNNs. It can achieve straightforward acceleration for on-device inference with the aid of compiler code generation, but it will obviously suffer from relatively high accuracy loss as whole kernel groups are pruned. On the other hand, the proposed KGS sparsity scheme is a more fine-grained structured sparsity that enjoys higher flexibility. In fact, the Vanilla sparsity scheme is just a special case of KGS sparsity, and therefore, one can confidently state that the KGS sparsity will result in a higher accuracy under the same pruning rate, as long as effective pruning algorithm has been developed and employed.


It is important to note that the KGS sparsity scheme is beyond a mere tradeoff of accuracy and mobile performance. In fact, with proper support of compiler code generation, the KGS sparsity can \emph{achieve almost the same mobile acceleration performance (e.g., in frames/second) as Vanilla sparsity}, under the same pruning rate. This is owing to the delicate design of KGS sparsity to match compiler-assisted mobile acceleration. For effective mobile acceleration, the whole kernel group will be transformed into matrix multiplication (with input feature map) \cite{chetlur2014cudnn} as shown in the reshaping step of Figure \ref{fig:KGSSparsity} (b). Accordingly, the KGS sparsity is equivalent to whole column removals in the weight matrix of a kernel group. The computation overhead in whole column removal is minor and can be mitigated by compilers, and the remaining computation is still based on full matrices (albeit smaller). A key observation is that the parallelism degree on off-the-shelf mobile devices is limited, and thus the smaller matrices of remaining weights \emph{have enough size to fully exploit the parallelism provided by mobile devices}. As an illustrative example, suppose that the mobile device can execute 10 operations in parallel while the matrix contains 100 remaining operations. Then the reduced-size matrix can be executed in 10 iterations, achieving full parallelism.
As the hardware parallelism can be fully exploited in both Vanilla and KGS schemes (if compiler overhead is negligible), the mobile acceleration performance in terms of FLOPs/second will be almost the same for both pruning schemes, and so does the frames/second performance under the same pruning rate (and FLOPs count). As a result, the proposed KGS sparsity can fully enjoy the benefit of high flexibility in terms of higher accuracy or higher pruning rate.

Please note that the $g_M \times g_N$ group size needs to be determined in Vanilla and KGS sparsity schemes, in order to achieve the maximum on-device parallelism with low computation overhead. The group size is determined offline with actual mobile testings using synthesized CNN layers. In other words, it will NOT become a hyperparameter in the pruning algorithm. $g_N=4$ and $g_M=4$ or $8$ are preferred to match the SIMD (Single Instruction, Multiple Data) parallelism provided by current mobile CPUs and GPUs. These values are large enough to exploit the on-device parallelism and small enough to provide enough pruning flexibility and accuracy, as shall be seen in the experimental results. 


\section{Structured Sparsity Learning Algorithms}





\begin{figure*}[hptb]
\centering
\includegraphics[width=.83\textwidth]{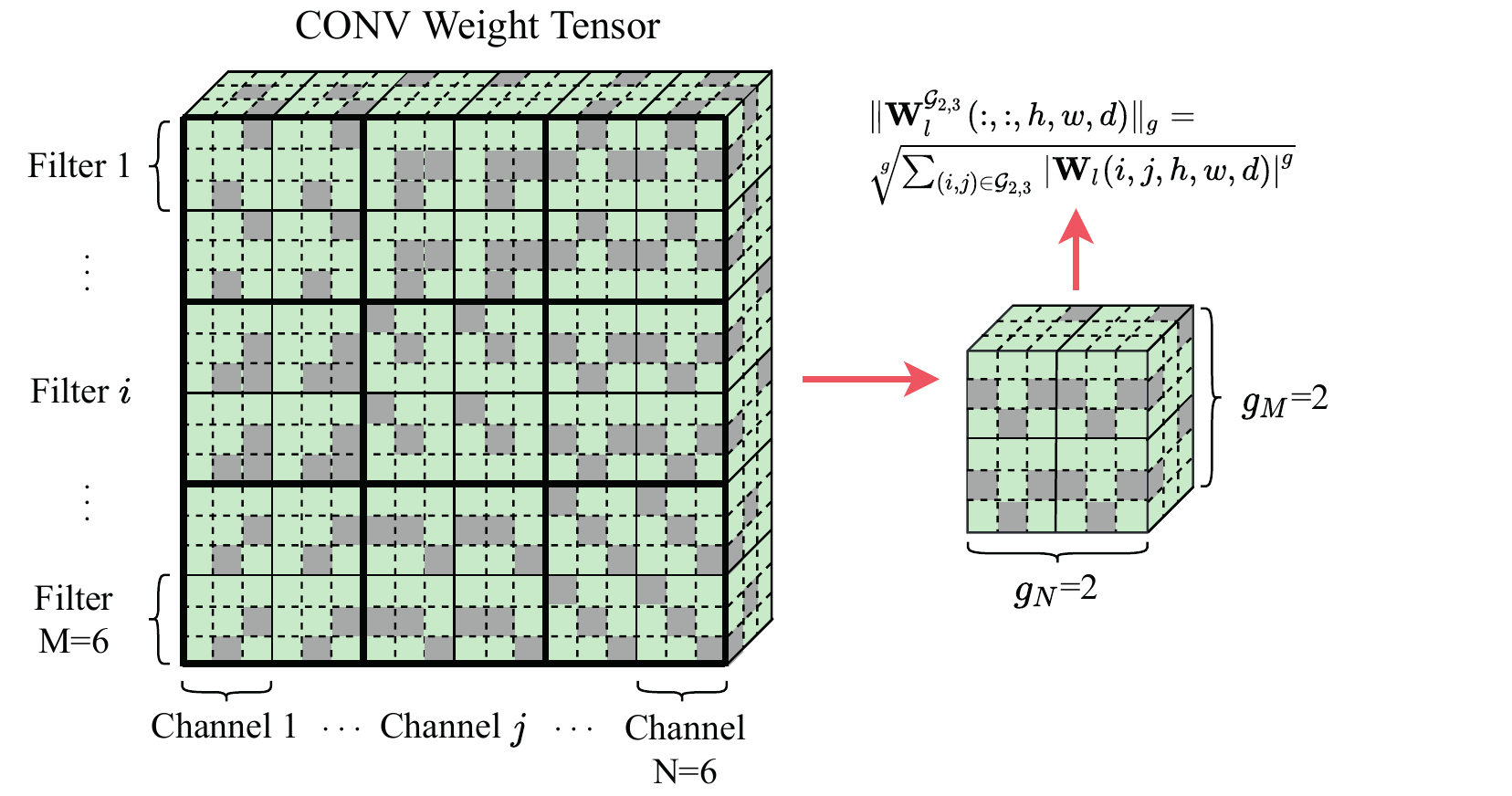}
\caption{An example of kernel groups, each consisting of $g_M\times g_N$ ($2\times 2$) 3D kernels. Within the same kernel group, kernels are pruned at the same locations (marked by the grey entries). To achieve the same sparsity pattern for kernels in the same group, group lasso is calculated as $\left\|\mathbf{W}_l^{\mathcal{G}_{p,q}}(:,:,h,w,d)\right\|_g=\sqrt[g]{\sum_{(i,j)\in\mathcal{G}_{p,q}}|\mathbf{W}_l(:,:,h,w,d)|^g}$.}
\label{fig:groupwisesparsity}
\end{figure*}

This section describes three pruning algorithms to achieve the proposed structured sparsity schemes for 3D CNNs. The first two are natural generalization from state-of-the-art algorithms on 2D CNN weight pruning, and the last one is specifically designed to address the limitations in the prior two.
Consider a general 3D CNN consisting of $L$ convolutional (CONV) layers.
Besides the $l$-th CONV layer weight tensor ${\mathbf{W}}_l$, the bias is denoted by ${\mathbf{b}}_l$.
The loss function associated with a 3D CNN can be denoted by $F(\{{\mathbf{W}}_l\}^L_{l=1},\{{\mathbf{b}}_l\}^L_{l=1})$.
To achieve the proposed group-wise sparsity schemes, weight tensor ${\mathbf{W}}_l$ is partitioned into a set of kernel groups along the dimensions of filters and channels, i.e., $\{{\mathbf{W}}_l^{\mathcal{G}_{p,q}}\}$, for $p \in [P]$ and $q \in [Q]$, where $P=\lceil M/g_M \rceil$, $Q=\lceil N/g_N \rceil$, and $[n]$ denotes the integer set $\{1,2,\ldots,n \}$.
Figure~\ref{fig:groupwisesparsity} provides an illustrative example of kernel groups.

\textbf{1. Heuristic Pruning Algorithm:} As discussed in Section~\ref{sec:relatedwork}, the prior work has investigated heuristic pruning for 2D CNNs, for both irregular and structured sparsity schemes. The prior work \cite{luo2017thinet,yu2018nisp} are mostly relevant to this work as we also focus on structured sparsity. Motivated by these work, we assign a similar “neuron importance score" to each kernel group (or the same location of kernels in the group), and perform pruning on the current layer with input information of the next layer in a back propagated manner (similar procedure as \cite{luo2017thinet}). This serves as our heuristic pruning algorithm for the proposed sparsity schemes of 3D CNNs.

\textbf{2. Regularization-based Pruning Algorithm:}
adds an additional regularization term to the loss function to achieve the Vanilla or KGS sparsity scheme.
Then, the regularization-based pruning can be formulated as
\begin{equation}\label{eqn:regularization_pruning}
\underset{ \{{\mathbf{W}}_{l}\},\{{\mathbf{b}}_{l} \} }{\text{minimize}}\ \  F \big( \{{\mathbf{W}}_l\}^L_{l=1},\{{\mathbf{b}}_l\}^L_{l=1} \big) + \lambda \sum^L_{l=1}R_g\big({\mathbf{W}}_l\big),
\end{equation}
where $R_g\big({\mathbf{W}}_l\big)$ is the regularization term for the Vanilla or KGS sparsity and $\lambda$ is the penalty measuring its importance.
Motivated by \emph{group Lasso} \cite{yuan2006model}, the regularization term can be defined as
$R_g\big({\mathbf{W}}_l\big) = \sum_{p=1}^{P}\sum_{q=1}^{Q} \left \|{\mathbf{W}}_l^{\mathcal{G}_{p,q}}\right \|_g$, where $\left \|\cdot\right \|_g$ denotes kernel group $\ell_p$ norm.
We can choose from $\ell_1$ norm {\cite{liu2017learning}}, $\ell_2$ norm {\cite{he2018soft, li2019compressing}} or their combination for this group-wise regularization.

In more details, the regularization-based pruning can be achieved by
\begin{equation}
\label{eqn:regularization_pruning_detail}
\begin{aligned}
& \underset{ \{{\mathbf{W}}_{l}\},\{{\mathbf{b}}_{l} \} }{\text{minimize}}\ F \big( \{{\mathbf{W}}_l\}^L_{l=1},\{{\mathbf{b}}_l\}^L_{l=1} \big) + \\
& \lambda \sum^L_{l=1}\sum_{p=1}^{P}\sum_{q=1}^{Q}\sum_{h=1}^{K_h}\sum_{w=1}^{K_w}\sum_{d=1}^{K_d} \left \|{\mathbf{W}}_l^{\mathcal{G}_{p,q}}{(:,:,h,w,d)}\right \|_g.
\end{aligned}
\end{equation}

\textbf{3. Reweighted Regularization Pruning Algorithm:} As discussed in Section~\ref{sec:relatedwork}, the fixed regularization-based pruning algorithm has a limitation, -- all weights will be equally penalized even after convergence of the pruning process, resulting in potential accuracy loss. We propose a novel reweighted regularization pruning algorithm to overcome this limitation. The basic idea is to systematically and dynamically reweight the penalties. Especially, we will reduce the penalties on weights with large magnitudes (which are likely to be more critical), and increase the penalties on weights with smaller magnitudes. This shall be performed in a systematic, gradual way to avoid the greedy solution which prunes a large number of weights at the early stage. Moreover, our proposed algorithm does not need to manually set the pruning rate for each layer, as a limitation in prior works based on ADMM or Geometric Median-based regularizations. 

For reweighted regularization, we minimize the following objective function:
\begin{equation}
\label{eqn:reweighted_regularization_pruning_detail}
\begin{aligned}
& \underset{ \{{\mathbf{W}}_{l}\},\{{\mathbf{b}}_{l} \} } {\text{minimize}}\ F \big( \{{\mathbf{W}}_l\}^L_{l=1},\{{\mathbf{b}}_l\}^L_{l=1} \big) + \lambda \bigg[ \sum^L_{l=1}\sum_{p=1}^{P}\sum_{q=1}^{Q} \\
& \sum_{h=1}^{K_h}\sum_{w=1}^{K_w}\sum_{d=1}^{K_d} \bigg(
{\mathcal P}_{l,t}^{\mathcal{G}_{p,q}} \circ \left \|{\mathbf{W}}_l^{\mathcal{G}_{p,q}}{(:,:,h,w,d)}\right \|_g \bigg) \bigg],
\end{aligned}
\end{equation}
where $\circ$ denotes element-wise multiplication. ${\mathcal P}_{l,t}^{\mathcal{G}_{p,q}}$ is the collection of penalty parameters and is updated in every iteration $t$ to facilitate the degree of sparsity. In each iteration, the instance of ${{\mathbf W}}_{l}^{\mathcal{G}_{p,q}}$ is denoted by ${{\mathbf W}}_{l,t}^{\mathcal{G}_{p,q}}$ and we update ${\mathcal P}_{l,t}^{\mathcal{G}_{p,q}}$ by setting 
\[
{\mathcal P}_{l,(t+1)}^{\mathcal{G}_{p,q}} = \frac{1}{\left \|{\mathbf{W}}_{l,t}^{\mathcal{G}_{p,q}}{(:,:,h,w,d)}\right \|_g^2+{\epsilon}},
\]
where $\epsilon$ is a small positive number avoiding the zero denominator. The reweighted regularization process updates penalty parameters based on the current weight values, will not incur extra hyperparameters, and has a fast convergence rate as analyzed in \cite{candes2008enhancing}. After 3$\sim$4 iterations, we will prune the weights that converge to zero, and perform a slight retraining on the non-zero weights (with a few epochs) to regain accuracy. 

While overcoming the limitation in fixed regularization-based algorithms, the advantage and flexibility in such algorithms will be preserved. There is only one $\lambda$ as the major hyperparameter, without the need of manually deciding per-layer pruning rate. Also similar to fixed regularization algorithms, we can multiply the per-layer FLOPs value to each layer $l$ in the above optimization function. In this way we can target at the overall FLOPs reduction, which is more relevant to the actual acceleration. In the experiments, we set the FLOPs reduction as the optimization target, and report the corresponding FLOPs reduction rates and actually measured mobile accelerations.

\section{Experimental Results}

\subsection{Evaluation on Sparsity Schemes and Pruning Algorithms}


\begin{table*}[hptb]
\centering
\begin{tabular}{|c|c|c|c|c|c|c|}
    \hline
    \multirow{2}{*}{Model} & Pruning & Sparsity & Overall FLOPs & Pruning Rate & Base Top-1 & Pruning Top-1 \\
                            & Algorithm & Scheme & after Pruning & of FLOPs & Accuracy & Accuracy \\
    \hline \hline
    \multirow{12}{*}{\makecell{C3D \\ (299MB)}} & \multirow{4}{*}{Heuristic} & Filter & 15.2G & 2.6$\times$ & \multirow{4}{*}{81.6\%} & 78.6\% \\
                            & ~ & Vanilla & 15.2G & 2.6$\times$ & ~ & 78.8\% \\
                            & ~ & KGS & 15.2G & 2.6$\times$ & ~ & 79.0\% \\
                            & ~ & KGS & 10.8G & 3.6$\times$ & ~ & 78.5\% \\ \cline{2-7}
                            
                            & \multirow{4}{*}{Regularization} & Filter & 15.2G & 2.6$\times$ & \multirow{4}{*}{81.6\%} & 78.8\% \\
                            & ~ & Vanilla & 15.2G & 2.6$\times$ & ~ & 79.0\% \\
                            & ~ & KGS & 15.2G & 2.6$\times$ & ~ & 79.6\% \\ 
                            & ~ & KGS & 10.8G & 3.6$\times$ & ~ & 79.3\% \\ \cline{2-7}
                            
                            & ~ & Filter & 15.2G & 2.6$\times$ & \multirow{4}{*}{81.6\%} & 79.3\% \\
                            & \textbf{Reweighted} & Vanilla & 15.2G & 2.6$\times$ & ~ & 79.7\% \\
                            & \textbf{Regularization} & \textbf{KGS} & \textbf{15.2G} & \textbf{2.6}$\times$ & ~ & \textbf{80.5}\% \\ 
                            & ~ & \textbf{KGS} & \textbf{10.8G} & \textbf{3.6}$\times$ & ~ & \textbf{80.2}\% \\ \hline \hline
    \multirow{12}{*}{\makecell{R(2+1)D \\ (120MB)}} & \multirow{4}{*}{Heuristic} & Filter & 15.9G & 2.6$\times$ & \multirow{4}{*}{94.0\%} & 89.0\% \\
                             & ~ & Vanilla & 15.9G & 2.6$\times$ & ~ & 89.4\% \\ 
                             & ~ & KGS & 15.9G & 2.6$\times$ & ~ & 90.4\% \\ 
                             & ~ & KGS & 12.7G & 3.2$\times$ & ~ & 89.9\% \\ \cline{2-7}
    
                             & \multirow{4}{*}{Regularization} & Filter & 15.9G & 2.6$\times$ & \multirow{4}{*}{94.0\%} & 89.8\% \\
                             & ~ & Vanilla & 15.9G & 2.6$\times$ & ~ & 90.8\% \\ 
                             & ~ & KGS & 15.9G & 2.6$\times$ & ~ & 91.7\% \\ 
                             & ~ & KGS & 12.7G & 3.2$\times$ & ~ & 91.3\% \\ \cline{2-7}

                             & ~ & Filter & 15.9G & 2.6$\times$ & \multirow{4}{*}{94.0\%} & 90.5\% \\
                             & \textbf{Reweighted} & Vanilla  & 15.9G & 2.6$\times$ & ~ & 91.7\% \\ 
                             & \textbf{Regularization} & \textbf{KGS} & \textbf{15.9G} & \textbf{2.6}$\times$ & ~ & \textbf{92.5}\% \\ 
                             & ~ & \textbf{KGS} & \textbf{12.7G} & \textbf{3.2}$\times$ & ~ & \textbf{92.0}\% \\ \hline
                             
\end{tabular}
\caption{3D CNN pruning results on the UCF101 dataset.}
\label{tab:prune_result}
\end{table*}

\textbf{Experimental Setup.}
We test the proposed two structured sparsity schemes i.e., Vanilla and KGS sparsity and three pruning algorithms on 3D CNN models (including one (2+1)D CNN): C3D~\cite{tran2015learning}, R(2+1)D~\cite{tran2018closer}, and S3D~\cite{xie2018rethinking}. 
Besides the two proposed sparsity schemes, a filter sparsity scheme is also implemented, where the filters may be pruned as a whole, and which is a direct generalization of the filter pruning of 2D CNNs.
The models are all pretrained on the Kinetics dataset~\cite{carreira2017quo} and transferred onto the UCF101~\cite{soomro2012ucf101} and HMDB51~\cite{kuehne2011hmdb} datasets as the pretrained dense models.
The hyperparameter settings are the same for all pruning algorithms and sparsity schemes for fair comparisons. 
The batch size is fixed to 32, and the video clip length is 16 frames. The initial learning rate is 5e$-$3 when training the dense model, and is reduced to 2e$-$4 in the weight pruning and retraining for stability. 
{The learning rate is fixed in the pruning process, while adjusted in retraining with a scheduler following the cosine function.}
For different types of sparsity schemes and pruning algorithms, the total number of epochs is fixed to 240 epochs.\footnote{Although the reweighted pruning algorithm is iterative, its latter iterations require significantly fewer epochs. Thus we can set the total epochs the same for different algorithms.}
For the pruning of all the models, we have used the best combination of $\ell_1$ and $\ell_2$ norms in the regularization term.
{The penalty factor $\lambda$ is set to 5e$-$4.}
The pruning and retraining processes are carried out with eight NVIDIA GeForce GTX 1080 Ti GPUs on Ubuntu operating system and the PyTorch~1.3 framework with CUDA~10.1. The total required memory is up to 38.8GB.

\paragraph{Results.}
The pruning results on C3D, R(2+1)D models on UCF101 dataset with various pruning algorithms and sparsity schemes are provided in Table \ref{tab:prune_result}.
For each pruning algorithm, the three sparsity schemes are compared under the same pruning rate (FLOPs reduction on the overall model), and KGS results of two pruning configurations are compared. As can be observed in the table, the KGS sparisity scheme consistently outperforms the vanilla sparsity, and these two schemes both perform better than filter pruning. The reweighted regularization algorithm consistently outperforms the other two pruning algorithms. The advantages of KGS sparsity and reweighted regularization are stated in Section 3 and Section 4. With reweighted regularization and KGS sparsity scheme, both C3D and R(2+1)D could achieve only 1\%$\sim$1.5\% accuracy loss under pruning rate of 2.6$\times$.

\subsection{Evaluation on Mobile Acceleration Performance}

\paragraph{Mobile Acceleration Framework Implementation.} We design and implement an end-to-end, compiler-assisted CNN acceleration framework that supports 3D CNNs. Without any pruning-related optimizations, \projectname is already faster than state-of-the-art CNN execution frameworks (such as MNN and PyTorch Mobile) on mobile CPUs, because we include more advanced optimizations like fine-tuned high-efficient SIMD (Single Instruction, Multiple Data) execution, fine-tuned weight layout organization, etc. Our framework is also the first to support 3D CNN executions on mobile GPUs. It is general, supporting both 2D and 3D CNNs. 
Comparing to other popular CNN acceleration frameworks that support 2D CONV (like TVM and MNN) on standard 2D benchmarks like VGG-Net, ResNet, MobileNet-V2, etc., our developed framework also yields consistently better performance. 

Moreover, \projectname is also the first to support various sparsity schemes, including Filter, and proposed Vanilla and KGS sparsity. Based on the sparsity scheme, it employs a compiler-based automatic code generation approach to reorganize the model weights, regularize the computations, tune the computation configuration, and generate the optimized model inference codes. Our framework can automatically generate both optimized CPU (vectorized C++) and GPU (OpenCL) codes to support both dense and sparse 3D CNN executions. 
\begin{table*}[hptb]
\centering
\begin{tabular}{|c|c|c|cc|cc|cc|cc|}
     \hline
     Framework & MNN & PyTorch & \multicolumn{4}{c|}{\projectname (Dense)} & \multicolumn{4}{c|}{\projectname (Sparse)} \\ \hline
     Device   & \makecell{CPU \\ (ms)} & \makecell{CPU \\ (ms)} & \makecell{CPU \\ (ms)} & Speedup & \makecell{GPU \\ (ms)} & Speedup & \makecell{CPU \\ (ms)} & Speedup & \makecell{GPU \\ (ms)} & Speedup \\ \hline
     \hline
     C3D      & 948 & 2544 & 902 & 2.8$\times$  & 488 & 5.2$\times$  & \textbf{357} & \textbf{7.1$\times$}  & \textbf{142} & \textbf{17.9$\times$}   \\ \hline
     R(2+1)D  & -   & 4104 & 1074 & 3.8$\times$ & 513 & 8.0$\times$  & \textbf{391} & \textbf{10.5$\times$} & \textbf{141} & \textbf{29.1$\times$}   \\ \hline
     S3D      & -   & 6617 & 1139 & 5.8$\times$ & 565 & 11.7$\times$ & \textbf{611} & \textbf{10.8$\times$} & \textbf{293} & \textbf{22.6$\times$} \\ \hline
\end{tabular}
\caption{Inference latency comparison of \projectname, MNN, and PyTorch on mobile CPU and GPU. 
MNN does not support R(2+1)D and S3D yet. For RT3D (Sparse), all models are pruned by reweighted regularization algorithm with KGS sparsity. The pruning rate (in FLOPs) is $3.6\times$ for C3D, $3.2 \times$ for R(2+1)D, and $2.1\times$ for S3D, and the accuracy is $80.2\%$, $92.0\%$, and $90.2\%$, respectively. }
\label{tab:performance-report}
\end{table*}

\begin{table*}[hptb]
\centering
\begin{tabular}{|c|c|c|c|c|c|cc|}
    \hline
    \multirow{2}{*}{Model}  & Sparsity & Base Top-1 & Pruning Top-1 & FLOPs & Pruning Rate & \multicolumn{2}{c|}{Latency (ms)}\\
                            & Scheme   & Accuracy   & Accuracy      & after Pruning & of FLOPs & CPU & GPU\\
    \hline \hline
    \multirow{2}{*}{C3D}    & Vanilla & \multirow{2}{*}{81.6\%} & \multirow{2}{*}{80.0\%} & 16.4G & 2.4$\times$ & 525 & 236 \\
                            & \textbf{KGS} & ~ & ~ & \textbf{9.7G}  & \textbf{4.0$\times$} & \textbf{329} & \textbf{134} \\ \hline\hline
    \multirow{2}{*}{R(2+1)D} & Vanilla & \multirow{2}{*}{94.0\%} & \multirow{2}{*}{91.8\%} & 15.5G & 2.5$\times$& 523 & 225 \\
                             & \textbf{KGS} & ~ & ~ & \textbf{10.2G} & \textbf{4.0$\times$}& \textbf{360} & \textbf{127} \\ \hline

\end{tabular}
\caption{Comparison between Vanilla and KGS sparsity schemes: pruning rate, and inference latency with the same pruning Top-1 accuracy on the UCF101 dataset. Reweighted regularization pruning is applied for all models.}
\label{tab:prune_ablation}
\end{table*}

\paragraph{Test-bed and Evaluation Setup.}
The evaluations are conducted on a Samsung Galaxy S20 cellphone with the latest Qualcomm Snapdragon 865 platform consisting of a Qualcomm Kryo 585 Octa-core CPU and a Qualcomm Adreno 650 GPU. 
All experiments run 50 times with 8 threads on mobile CPU, and all pipelines on mobile GPU. Because different runs do not vary severely, only the average inference execution time is reported for readability.
All models are tuned to their best configurations, e.g., with computational graph optimizations, the best tiling size, unrolling size, etc. 32-bit floating point is applied for CPU runs, and 16-bit floating point is used for GPU runs. This is the same for both baseline mobile acceleration frameworks and our \projectname framework for a fair comparison, as quantization is not supported by baseline frameworks.



\paragraph{Mobile Acceleration Results.} 
We next evaluate \projectname by comparing it with MNN~\cite{Ali-MNN} and PyTorch Mobile (PyTorch)~\cite{Pytorch-Mobile}.\footnote{Other popular mobile CNN acceleration frameworks like TVM and TFLite do not support 3D CNNs.}
Table~\ref{tab:performance-report} compares the end-to-end 3D CNN inference time (latency).
\projectname supports both dense (original) and sparse 3D CNNs on both mobile CPU and mobile GPU, PyTorch supports dense models on CPU only, and MNN supports dense C3D on CPU only.
For sparse models, \projectname uses pruned models by reweighted regularization pruning algorithms with KGS sparsity with the pruning rate of $3.6\times$ for C3D, $3.2\times$ for R(2+1)D, and $2.1\times$ for S3D, and the accuracy of $80.2\%$, $92.0\%$, and $90.2\%$\footnote{The base accuracy of S3D is 90.6\%.}, respectively.
In the table, the \projectname speedups are compared with PyTorch.
\projectname outperforms MNN and PyTorch on mobile CPU for all cases. \projectname on mobile GPU performs even better than on CPU. 
For example, for C3D, the fully optimized
\projectname (Sparse) outperforms the CPU version of PyTorch and MNN with the speedup of $7.1\times$ and $2.7\times$ on CPU, and $17.9\times$ and $6.7\times$ on GPU, respectively.
%
%
Notably, on mobile GPU, the fully optimized \projectname can infer 16 frames by using C3D, R(2+1)D, and S3D within 142 ms, 141 ms, and 293 ms, respectively, achieving real-time execution (say 30 frames per second) of 3D CNNs on mobile devices. 

Importantly, although \projectname's dense implementations have already been fully optimized, our sparse implementations especially on mobile GPU can fully transform the pruning rate of FLOPs into inference latency speedup. For example, on C3D, from \projectname (dense) to \projectname (sparse) on GPU, the improvement on inference latency is $3.43\times$, while the pruning rate of the sparse model is $3.6\times$.
This validates the statement in Section 3 that the proposed KGS sparsity scheme can exploit the parallelism degree on device. Moreover, 3D CONV is memory-intensive, bounded by both memory bandwidth and latency (which is more severe on mobile GPU due to its even limited cache capacity), and our pruning/compilation co-design is able to mitigate this issue with incurring negligible overhead. Our cache access count results validate this. 

\paragraph{Ablation Study.} We also compare two sparsity schemes, Vanilla and KGS in terms of pruning rate and inference latency on mobiles by controlling the same pruning top-1 accuracy (as shown in Table~\ref{tab:prune_ablation}). It shows that KGS scheme achieves both higher pruning rate (in FLOPs) and lower inference latency under the same pruning accuracy on both C3D and R(2+1)D due to KGS's high flexibility and seamless match with compiler-level optimizations.

\section{Conclusion}\label{sec:conclusion}

This paper presents \projectname, a mobile acceleration framework for 3D CNNs that includes two novel, mobile-friendly structured sparsity schemes (Vanilla and KGS) and best-suited pruning algorithms,
that can achieve low inference latency and high accuracy, simultaneously. Based on them, 
\projectname employs 
 a compiler-assisted code generation framework to transform pruning benefits to performance gains. The evaluation results show that \projectname beats two state-of-the-art acceleration frameworks with speedup up to $29.1\times$. \projectname can infer 16 video frames within 150 ms, for the first time, achieving real-time inference of 3D CNNs on off-the-shelf mobile devices with a pure software solution. 

\section*{Ethics Statement}
As a pure software solution, \projectname is the first to achieve real-time execution of 3D CNNs on mobile devices without notable accuracy loss--which can only be achieved by special (and more expensive) hardware solutions previously. This research will significantly encourage the general research of deep learning acceleration with software-based techniques while reducing the demand for some hardware-based accelerations. \projectname will enable many machine learning applications of behavior/activity detection on mobile platforms that have to run on the cloud previously.

The ethical aspects and future societal consequences of this research are highly application-dependent. This work has the following potential positive impact in society: First, because machine learning applications can run on the mobile (edge) side only without transmitting user data to the cloud server, data privacy is significantly enhanced, thus these applications can run in a more private environment. Second, this work may also significantly broaden the usage of machine learning in many other domains ranging from medical science, biology to geoscience and environmental science, e.g., combining motion sensors and high-precision 3D CNN inferences to support short-latency motion recognition can enable a real-time Parkinson treatment. 
At the same time, this work may have some negative consequences: due to the low-cost and easy-accessible nature of machine learning on mobile, this work has the potential of increasing the possibility of misusing machine learning techniques.
Furthermore, we should be cautious of the result of the failure of the system which could cause wrong decision making, thus jeopardizing the safety of the public and individuals. 
In addition, all experiments in our work are based on the public dataset and our task/method does not leverage biases in the data.

\section*{Acknowledgement}
This material is based upon work supported by the National Science Foundation (NSF) under Grants CCF-1937500, CNS-1932351, Army Research Office Young Investigator Program 76598CSYIP, and Jeffress Trust Awards in Interdisciplinary Research. Any opinions, findings, and conclusions or recommendations expressed in this material are those of the authors and do not necessarily reflect the views of NSF, or Thomas F. and Kate Miller Jeffress Memorial Trust. 

\nolinenumbers

\bibstyle{aaai21}
\bibliography{new_reference}

\clearpage  
\newpage  

\end{document}